\pdfoutput=1
\documentclass[11pt]{article}

\usepackage[review]{EMNLP2022}

\usepackage{times}
\usepackage{latexsym}

\usepackage[T1]{fontenc}
\usepackage[utf8]{inputenc}

\usepackage{microtype}

\usepackage{inconsolata}

\usepackage{hyperref}
\usepackage{longtable}
\usepackage{booktabs}
\usepackage{multirow}

\usepackage{microtype}
\usepackage{cite}
\usepackage{amsmath,amssymb,amsfonts}
\usepackage{algorithmic}
\usepackage{graphicx}
\usepackage{textcomp}
\usepackage{xcolor}
\usepackage{multirow}
\usepackage{subfigure}
\usepackage{mathrsfs}
\usepackage[ruled,vlined]{algorithm2e}
\usepackage{algorithm2e}
\usepackage{graphicx}
\usepackage{balance} 
\usepackage{color}
\usepackage{amsmath}
\usepackage{enumitem}
\usepackage{textcomp}
\usepackage{xcolor}
\usepackage{makecell}
\usepackage{bm}
\usepackage{booktabs}
\usepackage{tabu}
\usepackage{hyperref}
\usepackage{enumitem}
\usepackage{supertabular}
\usepackage{makecell}
\setenumerate[1]{itemsep=0pt,partopsep=0pt,parsep=\parskip,topsep=2pt}
\setitemize[1]{itemsep=0pt,partopsep=0pt,parsep=\parskip,topsep=2pt}
\setdescription{itemsep=0pt,partopsep=0pt,parsep=\parskip,topsep=2pt}
\newcommand{\sysname}{SENT\xspace}

\title{Learning to Detect Noisy Labels Using Model-Based Features}

\author{
    Zhihao Wang\thanks{$\;$ The authors have contributed equally to this work. The work was
conducted while the author was an intern at Recurrent AI.}$\;^{14}$, Zongyu Lin\footnotemark[1]$\;^2$, Peiqi Liu$^3$, Guidong Zheng$^3$, Junjie Wen$^3$\\
    \textbf{Xianxin Chen$^1$, Yujun Chen$^1$, Zhilin Yang\thanks{$\;$ Corresponding Author.}$\;^{12}$} \\
    $^1$Recurrent AI, $^2$Tsinghua University, $^3$China Merchants Bank, $^4$Meta\\
    \texttt{lucaswang@meta.com}, 
    \texttt{\{linzongy21, zhiliny\}@tsinghua.edu.cn}, \\
    \texttt{\{liupeiqi,zhengguidong,wenjunjieee}@cmbchina.com\} \\ % \texttt{jietang@tsinghua.edu.cn}
    \texttt{\{chenxianxin,chenyujun}@rcrai.com\}
}

\begin{document}

\maketitle
\begin{abstract}

Label noise is ubiquitous in various machine learning scenarios such as self-labeling with model predictions and erroneous data annotation. Many existing approaches are based on heuristics such as sample losses, which might not be flexible enough to achieve optimal solutions. Meta learning based methods address this issue by learning a data selection function, but can be hard to optimize. In light of these pros and cons, we propose \sysname (Selection-Enhanced Noisy label Training) that does not rely on meta learning while having the flexibility of being data-driven. \sysname transfers the noise distribution to a clean set and trains a model to distinguish noisy labels from clean ones using model-based features. Empirically, on a wide range of tasks including text classification and speech recognition, \sysname improves performance over strong baselines under the settings of self-training and label corruption.\footnotemark[1]

\end{abstract}

\footnotetext[1]{Code is available at \href{https://github.com/Rafa-zy/SENT}{https://github.com/Rafa-zy/SENT}}
\section{Introduction}

State-of-the-art deep neural networks require large amounts of annotated training data. Though the success of large pre-training models~\citep{devlin2018bert} alleviates such requirements, high-quality labeled data are still crucial to obtain the best performance on downstream tasks. However, it is extremely expensive to acquire large-scale annotated data for every new task.

To overcome the challenge of rigorous data requirements, recent works utilize weak labels for supervision, including heuristic rules ~\citep{augenstein2016stance,bach2019snorkel,awasthi2020learning}, large-scale datasets with cheap but noisy labels ~\citep{li2017webvision,lee2018cleannet};, and self-training ~\citep{park2020improved,wang2020adaptive}. In self-training, one trains a model on a labeled dataset and then predicts on a large amount of unlabeled data. Then the clean labeled data and pseudo-labeled data are combined to further train the model. 

The aforementioned sources of supervision share two important characteristics: First, they are learning from noisy labels. Second, the noise is dependent on data features. Thus, they could be unified into the framework of noisy label learning, for which numerous approaches have been proposed to reduce the negative impact of noise. However, there are three problems in prior work on noisy label learning: First, many existing approaches are based on heuristics such as sample losses which are not flexible enough ~\citep{han2018co,yu2019does,zhou2020robust,song2019selfie}; 
Second, many previous works require prior knowledge of the noise distribution of the dataset to adjust the hyperparameters, which is often not available in real-world applications ~\citep{song2019selfie,song2020learning}. Third, meta learning based methods avoid previous problems but suffer from optimization difficulties ~\citep{ren2018learning,zheng2021meta} such as longer training time, heavy hyper-parameter tuning and an unstable convergence process. To address the above problems, we propose a simple data-driven approach which does not rely on meta learning while being flexible.

% Empirically, we achieve better performance than  strong baselines. 
Our contributions are summarized as follows:
\begin{itemize}
    \item We propose a simple yet effective de-noising approach which avoids the optimization difficulty of meta learning while enjoying the flexibility of being data-driven.
    \item We unify the settings of both self-training and label corruption into a noisy label learning framework and demonstrate the effectiveness of our approach under both settings.
    \item Our approach improves performance over state-of-the-art baselines on a wide range of datasets, including text classification and automatic speech recognition (ASR).
    Last but not least, our approach achieves even larger gains on few-shot learning. 
    
    % our experiments show that our model can also achieve great gain under the setting of few-shot learning.
\end{itemize}

\section{Related Work}

\subsection{Self-training}
Self-training is a powerful learning method that enables models to learn from huge amounts of unlabeled data by generating weak labels through either the teacher model predictions or heuristic rules. Self-training has been shown to be effective in many scenarios, including image classification~\citep{yalniz2019billion}, text classification~\citep{li2019learning}, machine translation~\citep{wu2019exploiting}, etc. However, noise contained in weak labels could largely hinge the performance of self-training.

Recently,~\citet{xie2020self} improved the performance of self-training on image classfication by injecting noise to the student model, which is called NoisyStudent.~\citet{park2020improved} customized NoisyStudent on automatic speech recognition. One problem related with self-training is error propagation~\citep{zou2019confidence}; in other words, pseudo labelling on unlabeled data might bring noise to the training set which leads to the degradation of further training. Most previous work simply set a fixed threshold to filter samples with low confidence~\citep{sohn2020fixmatch,xie2020self}.~\citet{wang2020adaptive} used meta learning for adaptive sample re-weighting to mitigate error propagation from noisy pseudo-labels.~\citet{zhang2021flexmatch} used a curriculum learning approach to re-weight unlabeled data according to the model’s learning status. In our work, we alleviate the error propagation from another perspective, by learning a selection model using model-based features based on a clean dataset.

\subsection{Noisy Label Learning}
Learning from noisy labels has long been a research area. One of the most classical works is to add a noise adaptation layer on top of the main model to learn a label transition matrix for label correction~\citep{goldberger2017deep}. Bootstrapping~\citep{reed2014training} introduces the notion of perceptual consistency that a model predicts correct labels for noisy samples before overfitting to noisy labels. 
 Co-Teaching~\citep{han2018co} and Co-Teaching+~\citep{yu2019does} train  two networks while each network selects its small-loss samples as clean samples for its peer network.
However, the aforementioned approaches only deal with class dependent noise (CDN) and make a strong assumption that noise distribution is independent of each instance, which is not flexible enough for many cases.

%O2U-Net(\citet{huang2019o2u}) computed the statistics of losses of samples in different training stages to differentiate noisy and clean samples. 

SEAL~\citep{chen2020beyond} goes beyond previous work to consider instance dependent noise (IDN), which is more realistic and common than CDN on real world datasets. 
% SELF~\citep{nguyen2019self} leverages the strategies of both label refurbishment and sample selection by ensembling the snapshots of both models and model predictions to perform updates and filtering on training set.
SELFIE~\citep{song2019selfie} uses a hybrid approach that selects refurbishable samples based on the entropy of model predictions and then refurbishes the labels with model predictions. RoCL~\citep{zhou2020robust} utilizes curriculum learning that starts with easy and clean samples and gradually moves to  data with pseudo labels produced by a time-ensemble. However, both SELFIE and RoCL require prior knowledge of the noise distribution of the dataset and manual adjustment for hyperparameters. To avoid such efforts, meta learning is introduced to learn selection and refurbishment. 

Learning to Re-weight~\citep{ren2018learning} is a meta learning algorithm that learns to assign weights to training examples based on their gradient directions. Meta-weight-net \citep{shu2019meta} parameterizes the reweighting function as a multi-layer perceptron network. Meta Label Correction~\citep{zheng2021meta} trains the target model with corrected labels generated by a label correction model trained on clean validation data which is jointly trained by solving a bi-level optimization problem. 
 These meta learning algorithms afford a large degree of flexibility by directly optimizing a reliable objective. However, meta learning based models are known to be  sensitive to hyperparameter tuning and the quality of support data~\citep{agarwal2021sensitivity} and suffer from optimization difficulties  as they are trained by propagating second-order gradients~\citep{hospedales2020meta}. 

% Hence, we propose a simple yet effective algorithm - \sysname (\textbf{S}election \textbf{E}nhanced \textbf{N}oisy label \textbf{T}raining) using model-based features. We train a selection model on a small clean dataset and apply that model to distinguish clean samples from noisy datasets.
The major differences between our approach and previous methods are as follows.
Compared with meta-learning based models, our approach do not suffer from optimization difficulties. Compared with models with CDN assumptions, our approach can handle the IDN settings. Compared with other state-of-the-art IDN methods such as SELFIE and RoCL, the selection strategy in our approach is learnable with model-based features. And our approach does not require the prior knowledge of noise distribution. Last but not least, we unify the settings of self-training and label corruption in the framework of noisy label learning and conduct extensive experiments on both settings.

% \zy{Move part of the below sentence to the next section.}
% A wide range of experiment shows that our approach significantly excels in both self-training and noisy label learning.

\begin{figure*}[h]

     \centering
     \includegraphics[width=1.0\linewidth]{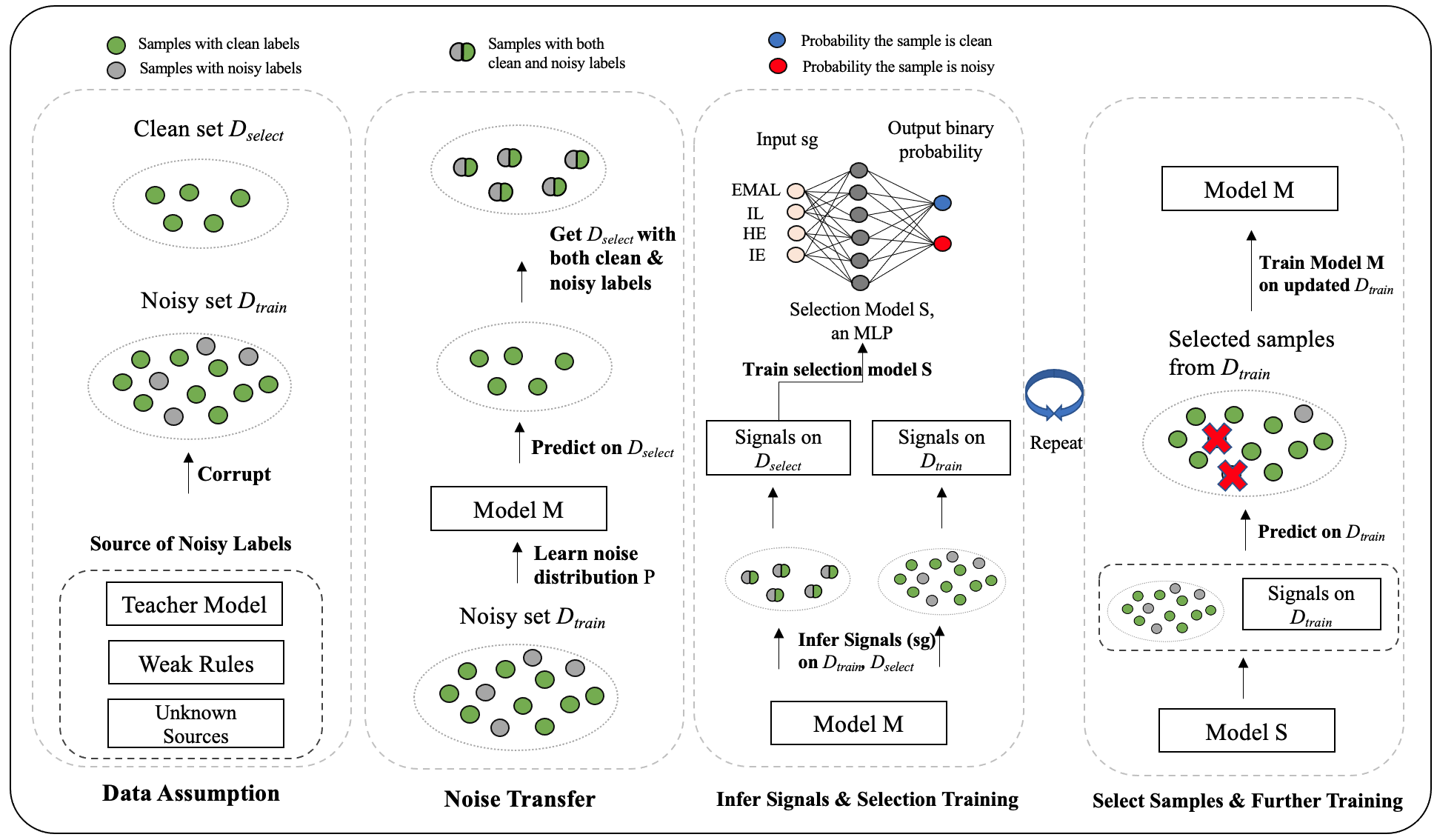}
     \caption{The pipeline of our general framework. Firstly, we are given a noisy training set and a small clean set for selection (also as development set). Secondly, we transfer the noise from the training set to the selection set. Thirdly, we compute predefined signals on both sets and train our selection model \textbf{S} using the selection set. Fourthly, We apply the trained selection model on the training set to distinguish clean samples for training model \textbf{M}. Finally, we repeat step 3 and 4.}
     \label{fig:system}
 \end{figure*}

\section{Method}

Now we present our approach, \sysname (Selection Enhanced Noisy label Training), that learns to select a subset from a  noisy dataset and only uses the selected subset for training to reduce label noise. The core idea is to transfer the noise distribution so that both clean and noisy labels are available on a data subset. A selection model is trained to distinguish clean labels from noisy ones and then applied to selecting a clean subset from a noisy dataset. 

Formally, given a noisy training dataset $D_{train} =  \{(x_i,y_i)| 1 \leq i \leq N\}$ that is corrupted following some unknown noise distribution $P$, our approach is to learn a selection strategy $f$ to select a clean subset $D'_{train}= \{x_i | y_i = y_i^*,i = 1,2,...,N\}$ for training. Here $(x_i,y_i)$ is a training sample, $y_i$ is the noisy label, $y_i^*$ is the corresponding unknown true label, and $N$ is the size of training set. Let model $M$ be our main model which is trained on the noisy dataset $D_{train}$ and performs certain tasks such as text classification, speech recognition and others. We will also have a selection model $S$ trained on a small clean development dataset $D_{select}$. The task of $S$ is to learn the selection strategy $f$. There are two main stages in our approach: noise transfer and selection learning.

\subsection{Noise Transfer}
Now we describe the noise transfer stage. Given the corrupted training set $D_{train}$, we learn the unknown noise distribution $P$ and transfer the noise to $D_{select}$.
We train a model $M$ on $D_{train}$ till full convergence, which predicts a (noisy) label given a text input. $M$ is now assumed to have learned to capture the unknown distribution $P$ by its parameters. Then we will use the model $M$ to get noisy labels on $D_{select}$ by making predictions. Here we argue that the noise on $D_{select}$ is approximately following the noise distribution $P$. 
Formally, now the development set is $D_{select} = \{ (x_{i}^{select},y_{i}^{select},y_{i}^{select*})\}$, where $(x_{i}^{select},y_{i}^{select*})$ is the original clean development sample, and $y_i^{select}$ is the noisy label predicted by $M$.
% , and $y_i^{select*}$ is corresponding clean label.

\subsection{Selection Learning}
\label{sec:SL}
We model the selection learning stage as a binary classification task. On $D_{select}$, a model $S$ is trained to classify whether a label is clean or noisy. In our approach, the model $S$ is constructed as a multi-layer perceptron with one hidden layer, which uses a pre-calculated 5-dimensional feature vector as the input and outputs a binary classification probability. Given a sample from the development set ${(x_{i}^{select},y_{i}^{select},y_{i}^{select*})}$, the selection training sample is defined as $({sg}_i,{sr}_i)$, where ${sg}_i$ is a 5-dimensional feature vector for the $i$-th sample. We will discuss how to compute the 5-dimensional feature in later sections. Meanwhile, ${sr}_i$ is the corresponding true selection result, defined as ${sr}_i = \mathbb{I}_{[{y_{i}^{select}=y_{i}^{select*}}]}$. In other words, if the $i$-th sample has a clean label, ${sr}_i = 1$, and otherwise zero. 
% As $S$ is modeling the selection strategy, we have $P_{select}(y_{i}^{select}|{sg}_i,S)$ as the predicted binary probability of the selection training sample ${sg}_i$. Then the loss function is defined as:
Let $P_{select}(sr_i|sg_i)$ denote the probability given by the selection model $S$. The loss function for each sample can be written as:
\begin{equation}
\label{equ1}
L_{i}^{select} = -\log P_{select}({sr}_i | sg_i)
\end{equation}

\subsection{Model-Based Features} \label{sec:features}

Now we discuss how to compute the 5-dimensional feature  $sg_i$ for each selection sample $x_i^{select}$.
% For each sample $x_{i}^{select}$, we are calculating a 5-dimensional feature ${sg}_i$, each dimension represents an informative signal. 

The first feature is called the instant loss. Given a sample ${(x_{i},y_{i})}$, let $\hat{P}(y_i | x_{i})$ be the predicted probability from the main model $M$. The instant loss (IL) for the sample is defined as:
\begin{equation}
\label{equ2}
\text{IL}_i = -log\hat{P}(y_i | x_i)
\end{equation}

Intuitively, a larger instant loss indicates a possibly noisier sample, because the model $M$ has low confidence in predicting the label. However, as training proceeds, the model will overfit some of the noisy labels. As a result, the instant loss will decrease for noisy samples as well.
To address this issue, following \citet{zhou2020robust}, we additionally use an exponential moving average (EMA) loss to better differentiate noisy and clean samples. In the $t$-th training epoch, the EMA loss (EMAL) for the $i$-th sample is defined as follows:
\begin{equation}
\small
\label{equ3}
\text{EMAL}_{i}^t = 
\left\{
    \begin{array}{lr}
    \gamma \text{EMAL}_{i}^{t-1} +  (1-\gamma) \text{IL}_i, & t \geq 1 \\
    \text{IL}_i, & t = 0 
    \end{array}
\right.
\end{equation}
where $\gamma \in [0,1]$ is a discounting factor. Intuitively, a larger EMAL represents a possibly noisier sample as the model has lower confidence in the training history.

% larger possibility that a sample is noisy as the model prediction of the sample is more different with the labels in the training history.

\citet{song2019selfie} has shown that the entropy of model prediction is a strong indicator to differentiate noisy and clean samples as noisy samples tend to have a larger entropy.
% because the model predictions of a noisy sample are unstable. 
Here we adopt two entropy signals as additional features: Instant Entropy (IE) and History Entropy (HE). The calculation of IE follows \citet{song2019selfie}. Let $\hat{y_i}^t$ be the predicted label of the $i$-th sample at epoch $t$ and let $H_{i}(t) = \{\hat{y_i}^0,\hat{y_i}^1,....,\hat{y_i}^{t}\}$ be the prediction history of first $t$ epochs. Then we formulate an empirical distribution
\[
\tilde{P}(y | x_i, t) = \frac{1}{t} \sum_{y' \in H_i(t)} \mathbb{I}_{[y = y']}
\]
which equals the ratio of prediction $y$ in the first $t$ epochs. The IH feature at the $t$-th epoch is computed as
\begin{equation}
\label{equ7}
\text{IE}_{i} = (1/\tau) \times - \tilde{P}(\hat{y}_i^t | x_i, t) \log \tilde{P}(\hat{y}_i^t | x_i, t)
\end{equation}
where $\tau = -\log(1/k)$ is a normalizing factor with $k$ being the number of labels. The HE feature at the $t$-th epoch is computed as
\begin{equation}
\label{equ5}
\text{HE}_{i} = (1/\tau) \times \sum_{y} - \tilde{P}(y|x_i, t) \log \tilde{P}(y|x_i, t)
\end{equation}

We explore another informative feature inspired by~\citep{han2019deep} who discovered that distances between low level features and high level features in a convolution model are larger on noisy samples than on clean samples. We find this holds for transformer models. Thus, we adopt the cosine similarity between the hidden states of the first layer and the last layer as another feature. Formally, FLS (First Last Similarity) is defined as:
\begin{equation}
\label{equ8}
\text{FLS}_{i} =\text{normalize} \left( \sum_j \cos(\mathbf{h}_{ij}^F, \mathbf{h}_{ij}^L) \right)
\end{equation}
where $\mathbf{h}_{ij}^F$ and $\mathbf{h}_{ij}^L$ represent the hidden states of the $j$-th token in the first and last layers respectively, and $\cos$ refers to cosine similarity. We normalize the FLS feature into the range of $[0,1]$.

% We assume that a smaller similarity indicates a larger possibility of a sample to be noisy.

Finally, we concatenate the above all features to be the input for the selection model $S$ as follows,
\begin{equation}
\label{equ9}
sg_i = [\text{EMAL}_{i},\text{IL}_{i},\text{HE}_{i},\text{IE}_{i}, \text{FLS}_i]
\end{equation}
where $[,]$ denotes concatenation.

\subsection{Overall Training Procedure}
\begin{algorithm}[htbp]
\small
	%\textsl{}\setstretch{1.8}
	\renewcommand{\algorithmicrequire}{\textbf{Input:}}
	\renewcommand{\algorithmicensure}{\textbf{Output:}}
	\caption{\sysname}
	\begin{algorithmic}[1]
		\STATE Initialization: $M$;$S$;$D_{train}$;$D_{select}$; total\_epochs; pretrain\_epochs
		\STATE Train $M$ on $D_{train}$. \COMMENT{Learn noise distribution $P$} \\
		\STATE Infer on $D_{select}$ using $M$. \COMMENT{Noise Transfer} \\
		\STATE epoch = 0; Reinitialize M;
		\STATE Pretrain $M$ for pretrain\_epochs;
		\WHILE{epoch $<$ total\_epochs} 
		  \STATE Infer signals $[\text{IL},\text{EMAL},\text{HE},\text{IE},\text{FLS}]$ on $D_{train}$ and $D_{select}$ using $M$;\\
		    \WHILE {not early stopping $S$}
		        \STATE Train $S$ on $D_{select}$;
		    \ENDWHILE
		    \STATE Select a clean subset $D'_{train}$ out of $D_{train}$ using $S$;
		  %  \STATE $P_{select}$ = $S(D_{train},sg_{train})$ 
		  %  \STATE $sr = argmax(P_{select})$
		  %  \STATE $D'_{train} = D_{train}[sr] $ \COMMENT{select clean train set using the prediction results of $S$}
		    \STATE Train $M$ on $D'_{train}$; 
		\ENDWHILE
		\ENSURE $M$
	\end{algorithmic}  
	\label{alg1}
\end{algorithm}

Now we show the training procedure with pseudo code in Algorithm~\ref{alg1} and illustrations in Figure~\ref{fig:system}. In the first box of Figure \ref{fig:system}, we clarify our data setting, where a clean set $D_{select}$ and a corrupted set $D_{train}$ are the input to our method.
% The difference in the settings of self-training and weak supervision of noisy label learning is that the sources of noises are different, which will be discussed in the following section.
% All settings require us to reduce the negative influence of noisy labels in the training set.
In the second box, we learn the noise distribution $P$ of the training set by fitting model $M$ on $D_{train}$ and then transfer $P$ to $D_{select}$. After this step, we will have both noisy and clean labels on $D_{select}$. This box corresponds to lines 2 to 4 in Algorithm~\ref{alg1}. 

Next, we will move on to the repeated training stage for model $S$ and model $M$, which is shown in box 3 and 4 of the figure. First, as illustrated in box 3, we use model $M$ to infer the aforementioned signals on $D_{train}$ and $D_{select}$. Then we will use the signals on $D_{select}$ as the input to train the selection model $S$. Since $D_{select}$ has both clean and noisy labels, we can easily get the training targets $sr$ as mentioned in Section~\ref{sec:SL} for the selection model $S$. Once we are done with selection learning, given the inferred signals on $D_{train}$, we can use $S$ to predict and select clean samples from $D_{train}$ to get $D'_{train}$. Then we will train $M$ on the clean subset $D'_{train}$ and repeat above steps. $D'_{train}$ is not guaranteed to be entirely clean but is expected to be cleaner than $D_{train}$. This repeated training phase corresponds to the code from lines 6 to lines 13 in Algorithm~\ref{alg1}. 
% One thing worth to notice is that during the training for $S$, we will set aside a small subset of $D_{select}$ to monitor the performance of $S$ in order to prevent overfitting.

\subsection{Adaptation to Self-Training}
Our above framework can be directly applied to learning scenarios with noisy labels. In this section, we will further discuss how to adapt this method to self-training, a classic semi-supervised learning paradigm.
Generally, it trains the teacher model on labeled data to infer pseudo labels on unlabeled data and add them back to the original training set. Then a student model is trained on the combined data. After that, the student becomes a new teacher model and the above process is repeated. Obviously, the process of pseudo labelling will bring noise since it cannot ensure 100\% accuracy on the predicted labels. Therefore, it fits the nature of our proposed framework. Specifically, the noise distribution $P$ in self-training is known because the source of noise is the teacher model. Thus, it is natural to directly leverage the teacher model to infer noisy labels on $D_{select}$.
% clean samples (dubbed as the noisy transfer process). 

\begin{figure}[h]
     \centering
     \includegraphics[width=0.75\linewidth]{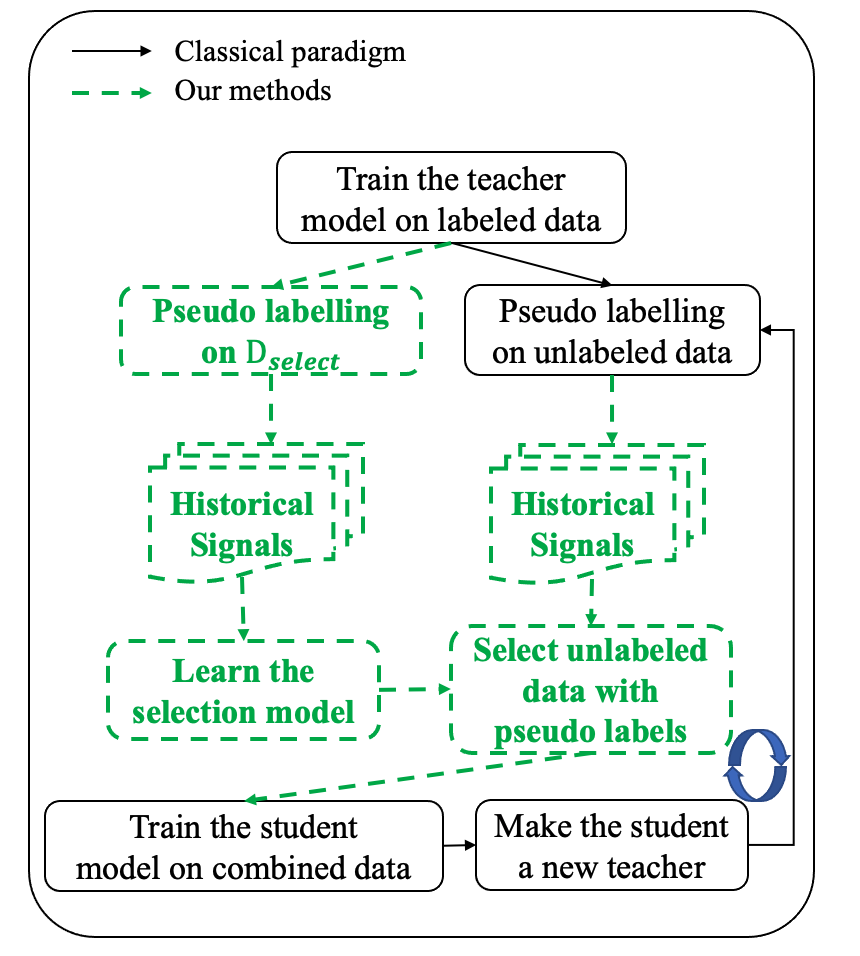}
     \caption{The pipeline of our general system applied to the classical self-training paradigm.}
     \label{fig:pipeline2}
 \end{figure} 

The whole pipeline of adapting our framework to classic self-training is displayed in Figure~\ref{fig:pipeline2}, and the corresponding algorithm is shown in Algorithm~\ref{alg2} in Appendix.
% At the initial stage, the dev data $D_{total}$ is split into $D_{select}$ and $D_{eval}$. 
After training the teacher model on labeled data $L$, we use the teacher model to infer on the unlabeled data $U$ and the dev set $D_{select}$. This is followed by training the selection model based on the signals of $D_{select}$. Then, we utilize the selection model to predict on the unlabeled data $U$ to judge whether to choose each pseudo-labeled sample or not. Then, we train the student model on the combination of original labeled data and selected samples from unlabeled data. The above procedure is repeated as in classical self-training.

\section{Experiments}

\begin{table*}[t]

  \centering
    \begin{tabular}{cccccccc}
    \toprule
          & \textbf{IMDB} & \textbf{SMS} & \textbf{SMS*} & \textbf{TREC} & \textbf{TREC*} & \textbf{YOUTUBE} & \textbf{AGNEWS} \\
    % \midrule
    \midrule
    Train & 250   & 69    & 61    & 68    & 60    & 77    & 60 \\
    Test  & 25000 & 500   & 392   & 500   & 500   & 392   & 7600 \\
    Unlabeled & 24500 & 4502  & 4948  & 4965  & 5409  & 1409  & 11876 \\
    Dev\_eval   & 126   & 250   & 31    & 251   & 30    & 40    & 32 \\
    Dev\_select & 124   & 250   & 31    & 249   & 32    & 38    & 32 \\
    \bottomrule
    \end{tabular}%
  \caption{Statistics of text classifcation datasets.}
  \label{tab:tcstats2_main_text}%
\end{table*}%

\begin{table*}[t]
% \small
  \centering
    \begin{tabular}{c|ccccccc}
    \toprule
       Method   & \textbf{TREC}  & \textbf{TREC*} & \textbf{SMS}   & \textbf{SMS*}  & \textbf{AGNEWS} & \textbf{IMDB}  & \multicolumn{1}{c}{\textbf{YOUTUBE}} \\
    % \midrule
    \midrule
    Supervised & 83.4  & 84.2  & 97.8  & 97.5  & 78.1  & 85.1  & 92.9  \\
    Co-teaching+ & 81.8  & 79.0  & 98.2  & 98.4  & 82.8  & 86.8  & 92.1  \\
    L2R   & 80.8  & 86.0  & 98.6  & 97.2  & 84.2  & 84.7  & 92.9  \\
    SELF  & 81.0  & 81.2  & 98.4  & 98.6  & 82.4  & 83.5  & 92.9  \\
    Self-train & 84.0  & 84.8  & 97.9  & 98.4  & 82.4  & 87.0  & 93.6  \\
    Self-train (thres) & 84.2  & 87.8  & 97.9  & 98.4  & 83.3  & 85.6  & 93.1  \\
    Noisy Student & \textbf{85.0}  & 88.6  & 98.4  & 99.0  & 85.0  & 87.7  & 93.9  \\
    \midrule
    Ours  & \textbf{85.0}  & 86.6  & \textbf{99.0}  & \textbf{99.2}  & 83.4  & 88.3  & 94.4  \\
    Ours + noisy & \textbf{85.0}  & \textbf{89.2}  & \textbf{99.0}  & \textbf{99.2}  & \textbf{86.0}  & \textbf{89.0}  & \textbf{95.2}  \\
    % \bottomrule
    \bottomrule
    \end{tabular}%
    \caption{Results for text classification in the self-training setting. We compare our approach with baselines under the BERT-based models. 
    % We use micro-F1 to evaluate the performance on SMS and accuraccy for the other datasets.
    }
  \label{tab:text2}%
\end{table*}%

\subsection{Experimental Setup}
\subsubsection{Overview}
To verify the effectiveness of \sysname, we conduct extensive experiments on text classification and automatic speech recognition (ASR) benchmarks. 

We use text classification tasks to evaluate the self-training setting. We perform finetuning based on the BERT \citep{devlin2018bert} model. We use ASR tasks to evaluate the label corruption setting. Our approach and other baselines are built on top of an encoder-decoder transformer network. Details of model configuration can be found in Appendix~\ref{sec:modeldt}.

\

\subsubsection{Datasets}
For text classification in the self-training setting, we evaluate our framework on the following five benchmark datasets: question classification TREC-6~\citep{li2002learning}, spam classification of SMS messages~\citep{almeida2011contributions}, spam classification of YouTube comments~\citep{alberto2015tubespam},
AG's news topic classification dataset ~\citep{zhang2015character}, and sentiment classification on IMDB movie reviews~\citep{maas-etal-2011-learning}.
Our data splits follow previous work \citep{karamanolakis-etal-2021-self}.
Related details are shown in Table~\ref{tab:tcstats2_main_text}. For the SMS and TREC datasets, we consider two separate versions. The datasets with $*$ have smaller development sets $D_{eval}$ and $D_{select}$ while the ones without $*$ have larger developments. Smaller development sets are more challenging for noisy label learning because selection learning has to perform on a smaller clean set. We use these two separate versions to test the robustness of our approach.

For ASR in the label corruption learning setting, we use AISHELL-1 ~\citep{bu2017aishell} as the benchmark. Following prior work, we model IDN using DNNs' prediction error 
~\citep{du2015modelling,menon2018learning}. Specifically, we train three small transformer models to corrupt the training set to different corruption levels: hard, medium, easy. The higher the error rate, the harder the corrupted dataset. In the following experiments of \sysname, the prior noise information (i.e. how the training set is corrupted) is assumed to be unknown. Related statistics are shown in Table~\ref{tab:asrstats} in Appendix.

% \subsubsection{Model Configuration}
% For the first setting of text classification, we utilize the same pre-trained word embeddings and the same teacher model (i.e., MLP) used in ASTRA.
% For the second setting of text classification, we utilize
% the popular pre-trained language models BERT as the initialization of our teacher and student models.
% Regarding the task of ASR, we employ an encoder-decoder transformer network. Details can be found in Appendix~\ref{sec:modeldt}.

\subsubsection{Evaluation}
For text classification, we report micro F1 for SMS and accuracy for the rest of the datasets.
% the classification accuracy for text classification tasks.
% For the first setting of text classification, we report the same evaluation metrics and the results of the baselines used in ASTRA. Note that we also train our model for five different random splits of the labeled data and evaluate on the held-out test data. For the second setting of text classification, we report the above evaluation metrics on the corresponding datasets based on the BERT pre-trained model.
For ASR, we follow~\citet{bu2017aishell} to use the character error rate (CER) for evaluation.
% We show CER on the official test set in following Table~\ref{tab:asrres}. CER on both development set and testing set will be reported in appendix.

Since our approach relies on an additional clean set $D_{select}$, we split the normal development set into two halves. We use one half as $D_{select}$ for selection learning and the other half $D_{eval}$ for standard model tuning, so as to set up fair comparison with the baselines.

\subsubsection{Baselines}

For text classification, we compare with the following baselines: (a) ``Supervised'' refers to supervised learning using only labeled data; (b) ``self-train'' is standard self training that utilizes both labeled and unlabeled data for iterative training; (c) ``self-train (thres)'' means self-training that uses the development set to select a threshold of confidence score for filtering pseudo-labeled data; (d) ``noisy student''~\citep{xie2020self} adds dropout noise to the student model in self training; (e) ``co-teaching+''~\citep{yu2019does} uses two neural networks to select small-loss samples for each other and applies a disagreement strategy; (f) ``L2R''~\citep{ren2018learning} learns to re-weight noisy labels via meta-learning (g) ``SELF''~\citep{nguyen2019self} utilizes self-ensemble predictions to progressively remove noisy labels. We also evaluate the performance when combining noisy student and our method, denoted as ``ours + noisy''.

In the experiments of ASR, we include the following baselines: Vanilla (a naive encoder-decoder transformer network), Co-Teaching+ \citep{yu2019does},  L2R \citep{ren2018learning}, RoCL \citep{zhou2020robust} and SELFIE \citep{song2019selfie}. The details of these baselines can be found in Appendix~\ref{sec:baselines}.

\subsection{Experimental Results}

\begin{table}[t]
% \small
  \centering
    \begin{tabular}{c|ccc}
    \midrule
    Model & Hard  & Medium & Easy  \\
    % \midrule
    \midrule
    Vanilla & 32.63  & 21.82 & 16.10      \\
    Co-Teaching+ &  32.08 & 21.67 & 15.11  \\
    L2R &  30.43  & 20.07  & 15.15  \\
    RoCL &  27.16 & 17.87 & 14.95      \\
    SELFIE & 27.31 & 19.10 & 13.98 \\
    Ours & \textbf{26.91} & \textbf{17.67} & \textbf{13.47}       \\

    \bottomrule
    \end{tabular}%
  \caption{Comparison with baselines on AISHELL-1 test set in the label corruption setting. We use CER as the metric of performance. A lower CER indicates a better model.}
  \label{tab:asrres}%
\end{table}%

\begin{table}[t]
  \centering
    \resizebox{\linewidth}{!}{\begin{tabular}{c|c|cccc}
    \toprule
    Data  & Method & Pse-Acc. & Sel-Pre. & Sel-Rec. & \#Selected \\
    \midrule
    \multirow{3}[2]{*}{IMDB} & Self-train & 85.7  & 85.7  & \textbf{100.0}  & 24500 \\
          & Self-train (thres) & 85.2  & 94.3  & 58.1  & 12870 \\
          & Ours  & \textbf{88.1}  & \textbf{97.9}  & 47.5  & 7742 \\
    \midrule
    \multirow{3}[2]{*}{YOUTUBE} & Self-train & 95.0  & 95.0  & \textbf{100.0}  & 1409 \\
          & Self-train (thres) & \textbf{95.7}  & 96.1  & 98.7  & 1386 \\
          & Ours  & 94.6  & \textbf{96.9}  & 94.1  & 1294 \\
    \midrule
    \multirow{3}[2]{*}{SMS*} & Self-train & 98.1  & 98.1  & 100.0  & 4948 \\
          & Self-train (thres) & 96.0  & 96.0  & 98.6  & 4880 \\
          & Ours  & \textbf{98.3}  & \textbf{98.3}  & \textbf{100.0}  & 4897 \\
    \midrule
    \multirow{3}[2]{*}{TREC*} & Self-train & 77.4  & 77.4  & \textbf{100.0}  & 4965 \\
          & Self-train (thres) & 77.2  & 77.2  & 99.8  & 4944 \\
          & Ours  & \textbf{78.3}  & \textbf{81.2}  & 67.1  & 4897 \\
    \midrule
    \multirow{3}[2]{*}{AGNEWS} & Self-train & \textbf{84.4}  & 84.4  & \textbf{100.0}  & 11876 \\
          & Self-train (thres) & 83.4  & 83.6  & 99.7  & 11826 \\
          & Ours  & 84.3  & \textbf{88.0}  & 85.6  & 9728 \\
    \bottomrule
    \end{tabular}}%
      \caption{The key metrics of pseudo labeling and sample selection during the self-training. We report the accuracy of pseudo labeling, the precision and recall of sample selection, and the number of selected samples.}
  \label{tab:keymetrics}%
\end{table}%

\subsubsection{Results in Text Classification.}

% Since pre-trained language models significantly boost the performance on downstream tasks, we experiment with our approach on top of BERT. 
As shown in Table~\ref{tab:text2}, the self-training baseline improves text classification performance.
% applying naive self-train on BERT could lift the performance. 
%However, if we apply self-train and add a confidence threshold to filter the selected samples, the performance is not guaranteed to be better. 
In comparison, our selection approach can stably lift the performance, which shows that our selection model has learned an informative selection strategy. Last, although using \sysname alone outperforms self-train and noisy student, our approach can be combined with the noisy student approach to achieve even better performance. Overall, this combined approach achieves the best performance among all the datasets we consider.

\subsubsection{Results in ASR}
 Table~\ref{tab:asrres} shows that Co-Teaching+ is not able to handle our setting as it only achieves similar result to the vanilla model. L2R is effective on our problem setting with improved performance. However, meta learning based L2R underperforms RoCL and SELFIE. 
%  RoCL and SELFIE achieve superior performance after searching hyperparameters with the prior knowledge of noise distribution (such as noise rates). 
Compared with above baselines, our approach consistently excels on all error levels, which demonstrates the effectiveness of our approach.

% All baselines are reproduced with grid hyperparameter searching. *When searching prarameter for RoCL and SELFIE, prior knowledge about the noise distribution is used.
\subsection{Empirical Analysis}

\paragraph{Case Study: Key Metrics During Self-Training}
In order to gain a deeper understanding of how our method improves over the traditional self-training methods, we investigate some of the key metrics regarding pseudo labelling and selection performance. We consider self-train, self-train (thres), and our model. Specifically, we display the accuracy of the pseudo labelling on unlabeled data (Pse-Acc.), the precision of sample selection (Sel-Pre.), the recall of sample selection (Sel-Rec.) and the number of selected samples in the best round (i.e., the training round that achieves the best performance in the repeated self-training process). As can be seen in Table~\ref{tab:keymetrics}, our approach achieves a better pseudo labeling accuracy. This is because our approach obtains a more balanced tradeoff between selection precision and selection recall compared to self training. Because of a more rigorous selection model, our approach tends to only select samples with a higher probability of having clean labels; i.e., increasing the selection precision. This is also reflected in small decrease in the number of selected samples. We believe this is crucial for mitigating error propagation \citep{xie2020self} and thus for better performance.

\paragraph{Ablation Study: Substituting with Simpler Models}
To further test the robustness of our approach, we substitute the base model from BERT to simple multi-layer percetrons (MLPs) without pretraining. As show in Table~\ref{tab:text1}, the performance will decrease after applying simpler MLPs. However, our approach remains effective compared to the other baselines. The relative gain and absolute improvement from ``Supervised'' to our approach is still significant.
% even larger than BERT-based models.
\begin{table}[t]
\small
  \centering
    \resizebox{\linewidth}{!}{\begin{tabular}{c|cccc}
    \toprule
     Method     & TREC  & SMS   &  \multicolumn{1}{c}{YOUTUBE} \\
    % \midrule
    \midrule
    Supervised & 66.5 & 93.3 & 91.0 \\
    Co-teaching+ & 66.8 & 98.3 & 93.3  \\
    L2R   & 66.4 & 98.0 & 93.3  \\
    SELF  & 66.3 & 98.1 & 92.3  \\
    Self-train & \textbf{71.1} & 95.1 & 92.5  \\
    Self-train (thres) & 70.1 & 98.1 & 92.1  \\
    Noisy Student & 68.9 & 98.1 & 92.1  \\
    \midrule
    Ours  & 70.3 & 98.2 & 93.3  \\
    Ours+Noisy & 70.0 & \textbf{98.4} & \textbf{93.4} \\
    % \bottomrule
    \bottomrule
    \end{tabular}}%
    \caption{Comparison with baselines under the MLP models.}
  \label{tab:text1}%
\end{table}%

\begin{table}[t]
% \small
  \centering
    \begin{tabular}{c|ccc}
    
    \midrule
    Signal & Hard  & Medium & Easy  \\
    % \midrule
    \midrule
    All & \textbf{26.91}  & \textbf{17.67} & \textbf{13.47} \\
    -EMAL       & 27.31  & 17.95 & 13.85  \\
    -IL        & 28.56  & 19.32 & 14.94  \\
    -HE             & 29.32  & 19.88 & 15.01 \\

    \bottomrule
    \end{tabular}%
    \caption{Ablation experiments for features on AISHELL-1. 'All' is all features. Each line removes one feature based on the last line.}
    \label{tab:signalabalationonasr}%
\end{table}%

\begin{table}[t]
\scriptsize
  \centering
    \begin{tabular}{c|ccccccc}
    \toprule
    Signal & IMDB  & YT & TREC  & TREC* & SMS   & SMS*  & AG \\
    % \midrule
    \midrule
    All & 85.9 & \textbf{95.2}  & \textbf{85.0}  & \textbf{89.2}  & \textbf{99.0}  & \textbf{99.2}  & 85.1  \\
    -FLS & 88.2  & 92.9  & 83.6  & 82.0  & 99.0  & 99.0  & 84.1  \\
    -EMAL & \textbf{89.0}  & 92.1  & 84.2  & 86.8  & 99.0  & 99.0  & \textbf{86.0}  \\
    -IL & 87.8  & 92.6  & 83.4  & 85.8  & 99.0  & 98.8  & 82.5  \\
    -HE    & 88.0  & 93.4  & 82.6  & 79.8  & 98.4  & 98.0  & 79.6  \\

    \bottomrule
    \end{tabular}%
    \caption{Ablation experiments for features on text classification. 'All' is all signals. Each line removes one feature from the previous line. YT represents YouTube dataset, and AG represents AG News dataset.}
  \label{tab:signalabalationontc}%
\end{table}%

\begin{figure}[t]
     \centering
     \includegraphics[width=0.8\linewidth]{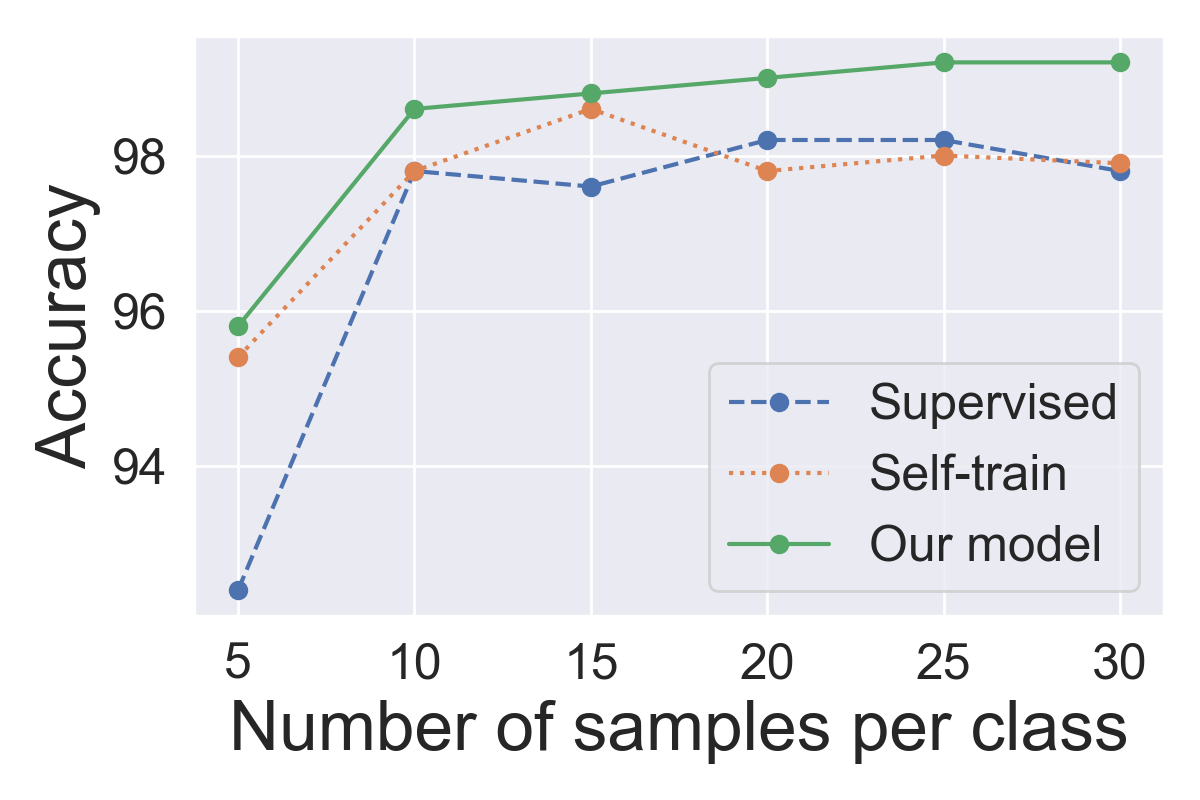}
     \caption{Evaluation on SMS under the few-shot setting.}
     \label{fig:few_sms}
 \end{figure} 
 
\paragraph{Ablation Study: Features}

We study the effects of the model-based features we introduced in Section \ref{sec:features} with an ablation. Note that we did not use FLS for ASR because the the first layer and the last layer have different lengths.
% We check that each signal is helpful for the selection learning on ASR. 
As shown in Table~\ref{tab:signalabalationonasr} and Table \ref{tab:signalabalationontc}, all of the features contribute to the final performance. Among the features, adding the instant loss (IL) feature results in the most relative gain for performance. 
% One thing worth to notice is that we do not use FLS on ASR task because the length of hidden states in decoders is different than the length of hidden states in encoders on ASR task.

Also, we do similar experiments on text classification tasks. As seen in Table~\ref{tab:signalabalationontc}, we can do basic search on feature engineering to improve the final performance. Among all signals, IL leads to the greatest relative gain of the performance. But across all the datasets, each signal has its own role and contributes to our final performance.

\paragraph{Performance Under Few-shot Setting}
LST~\citep{li2019learning} has shown that self-training paradigm can be customized on few-shot classification. Here, we also investigate the effectiveness of our method when applying to the few-shot setting. Specifically, we evaluate on selected text classification datasets (i.e., YOUTUBE, SMS and IMDB). Figure~\ref{fig:few_shot} showcases that generally self-train performs better than ``Supervised'' (using only labeled data), while our model achieves the best performance in most cases, indicating the robustness of our method. More results are displayed in Appendix~\ref{sec:few_shot}.

\section{Conclusions}
In this paper, we propose \sysname to address the problem of label noises. Compared with meta learning based models, our selection model is trained with full supervision using cross entropy loss which facilitates the convergence process. In the meantime, we model IDN noise without the prior knowledge of noise distribution. We also unify the setting of self-training and label corruption in the framework of noisy label learning and conduct extensive experiments on both settings.

\section{Limitations}
Although our framework has been proved to be effective under the setting of self-training and label corruption on text classification and speech recognition tasks, adapting our approach to more sequence-level tasks such as named entity recognition (NER) and machine translation will also be interesting. Besides, the selection model in our framework is feature-based. These features are very informative but might be limited in terms of expressivity. This reserves the space for further improvement to learn more data-driven features under our framework.

% \newpage
\bibliography{emnlp}
\bibliographystyle{acl_natbib}

\newpage

\appendix
\section{Appendix}
\label{sec:appendix}
% implementation (Section A.1)
We provide details of our datasets (Section A.1) and experimental results (Section A.2). 
% Our code is available at ...

\begin{table*}[htbp]
% \small
  \centering
    \begin{tabular}{ccccccc}
    \toprule
    Stats & OriginalTrain & HardTrain  & MediumTrain & EasyTrain & OriginalDev & OriginalTest  \\
    \midrule
    \midrule
    CER  & $NA$  & 37.11  & 26.19 & 16.14 & $NA$ & $NA$    \\
    MaxLen & 44  & 82  & 89 & 44 & 35 & 37 \\
    AvgLen & 14.41  & 14.30  & 14.35 & 14.41& 14.33 & 14.60     \\
    Num & 120098  & 120098  & 120098 & 120098  & 14326 & 7176      \\
    \bottomrule
    \end{tabular}%
    \caption{Statistics of AISHELL-1. ‘Worse','Normal','Better' are three notation we use to represent different error levels. 'NA' means not char error rate because original datasets are assumed to be clean. ’Num' represents the data volume of each set.}
  \label{tab:asrstats}%

\end{table*}%

\subsection{Details of Implementation}
% \subsubsection{Algorithm for applying our method to self-training}
 \begin{algorithm}[htbp]
 \small
	%\textsl{}\setstretch{1.8}
	\renewcommand{\algorithmicrequire}{\textbf{Input:}}
	\renewcommand{\algorithmicensure}{\textbf{Output:}}
	\caption{Adaptation of \sysname on classical self-training.}

	\begin{algorithmic}[1]
		\STATE Initialization:Labeled data $L$, Unlabeled data $U$, total dev data $D_{total}$, dev\_select data $D_{select}$;
		\STATE Initialize selected\_train\_set $L_{chosen}= L$;
		\STATE Initialize the teacher model $M_{teacher}$, selection model $S$;\\
		\STATE Train the teacher model $M_{teacher}$ on $L_{chosen}$; \\
		\STATE Infer on $U, D_{select}$ using $M_{teacher}$; %\COMMENT{The pseudo labelling causes label noise.} \\
        \STATE Infer signals $sg$ = $[\text{IL},\text{EMAL},\text{HE},\text{IE},\text{FLS}]$ on $U,D_{select}$ using $M_{teacher}$ and get signals $sg_{unlabeled},sg_{select};$ %for these three data respectively;\\	
        % \WHILE {not early stopping $S$}
		\STATE Train $S$ on $D_{select}$; 
% 		      %  \STATE Evaluate $S$ on $D_{eval}$;
% 		\ENDWHILE
	    \STATE $P_{select}$ = $S(U,sg_{unlabeled})$;  %\COMMENT{Use the selection model to infer on unlabeled data with pseudo labels.}
	    \STATE $sr = argmax(P_{select})$; %\COMMENT{Convert the probability into choose-or-not variables.}
	    \STATE $L_{chosen} = L \cup U[sr] $; %\COMMENT{Com  and add pseudo-labeled data original train data.}
	    \STATE Train the student model $M_{student}$ on $L_{chosen}$;
	    \STATE Make the student a teacher and go back to step 5 and repeat. 
	\end{algorithmic}  
	\label{alg2}
\end{algorithm}
\subsubsection{Baselines}
\label{sec:baselines}
In the first experiment of text classification, we also take the same baselines which consider rules and report the same results as utilized in ASTRA (~\citet{karamanolakis-etal-2021-self}). (a) Majority predicts the majority vote of the rules with ties resolved by predicting a random class. (b) Snorkel+Labeled (\citet{ratner2017snorkel}) trains
classifiers using weakly-labeled data with a generative model. . (c) L2R (\citet{ren2018learning}) learns to re-weight noisy labels from rules by meta learning. (d) PosteriorReg (\citet{hu2016harnessing}) uses rules as soft constraints for training by posterior regularization. (e) ImplyLoss (\citet{awasthi2020learning}) learns both instance-specific and rule specific weights by minimizing an implication loss (h) ASTRA (\citet{karamanolakis-etal-2021-self})
introduces an rule attention network to leverage multiple sources of weak supervision with trainable weights to compute soft weak labels for unlabeled data.

For ASR baselines, Vanilla is a naive encoder-decoder transformer network without any denoising moduels. All following baselines and our approache are build on top of the vanilla model. (b) Co-Teaching+ trains two networks with each network selecting its small-loss samples as clean samples for its peer network. (c) L2R is the same as mentioned before. (d) RoCL starts learning with easy and clean samples and gradually moves to learn noisy-labeled data with pseudo labels produced by a time-ensemble of the model and data augmentations.(e) SELFIE selects refurbishable samples based on the entropy of model predictions and refurbs the samples with model predictions.

\subsection{Details of Experiments}
\label{sec:modeldt}
For ASR models, the transformer model contains 12 layers of encoder and 6 layers of decoder. For each transformer block, the number of heads in the multiheadattention module is 4. The dimension of the encoder and decoder input is 256. The dimension of the feedforward network is 2048. We use 80-dimensional filter bank coefficient as input features.
The hyper parameter for training is show in Table ~\ref{tab:hpasr}. Batch\_size\_in\_s2 means the maximum allowed length of audio in seconds in one batch. History\_length represents the maximum allowed length for stored history predicted labels. These history predictions are used to calculate entropy. 

It should be noted that in both text classification as ASR tasks, we split the total development set into dev\_select and dev\_eval, where dev\_select is used to train the selection model and dev\_eval is used to evaluate the selection model.

\begin{table}[h]
\small
  \centering
    \begin{tabular}{cccc}
    \toprule
    HP & HardTrain  & MediumTrain & EasyTrain \\
    \midrule
    \midrule
    Max\_LR  & 5e-4  & 3e-4  & 3e-4    \\
    Min\_LR & 5e-6  & 1e-6  & 1e-6 \\
    Warmup\_step & 20000 & 20000 & 20000 \\
    Max\_steps & 80000  & 50000 & 50000     \\
    Batch\_size & 300  & 300 & 300      \\
    Batch\_size\_in\_s2 & 500 & 500 & 500 \\
    History\_length & 18 & 12 & 12 \\
    \bottomrule
    \end{tabular}%
    \caption{Hyperparameter for \sysname on ASR. HP means hyperparameters.}
  \label{tab:hpasr}%

\end{table}%

There are three details worth to noice in ASR: a) we perform an additional correction module for ASR. The correction module has the same architecture as the selection module. Correction module takes the same signal as selection module, and it outputs three weights which sum to one. The weights are assigned to the noisy labels (NL), model predicted probabilities (Pred), accumulated corrected labels respectively. The corrected label(CL) at T-th epoch is:
\begin{equation}
\label{equ10}
CL^T = w_1 * NL + w_2 * Pred + w_3 * CL^{T-1}
\end{equation}

after correction, we will perform the normal selection module to select clean labels from corrected labels. b) Since ASR is sequence level problem, we can not correct and select each token independently which would ignore the word dependencies. We first align the predicted word sequence to the noisy target sequence accoding to their longest common sequence. Then we will correct and select the token that are not common in both sequence. c) As ASR is a generation problem and the length of input and output is different, we do not extract FSL as a feature for our approach.

\subsection{Details of Ablation Study}
 \begin{table}[tbp]
% \small
  \centering
    \begin{tabular}{c|ccc}
    \toprule
    Model & Hard  & Medium & Easy  \\
    \midrule
    \midrule
    Vanilla &  30.06    &  20.21     &  14.42    \\
    Co-Teaching+ & 29.67 & 20.02 & 13.71  \\
    L2R & 27.79  & 19.08 & 14.12   \\
    RoCL & 24.50   & 16.34      & 13.22       \\
    Selfie & 24.40 & 17.22    & 13.01   \\
    ours & \textbf{23.97}     &  \textbf{16.01}     & \textbf{11.96}     \\

    \bottomrule
    \end{tabular}%
  \caption{Comparison with other baselines on dev set of AISHELL-1}
  \label{tab:asrresfull}%
\end{table}%
\subsubsection{The Influence of Selection Threshold On The Final Performance}
\begin{figure}[htbp]
\centering
\includegraphics[width=0.8\linewidth]{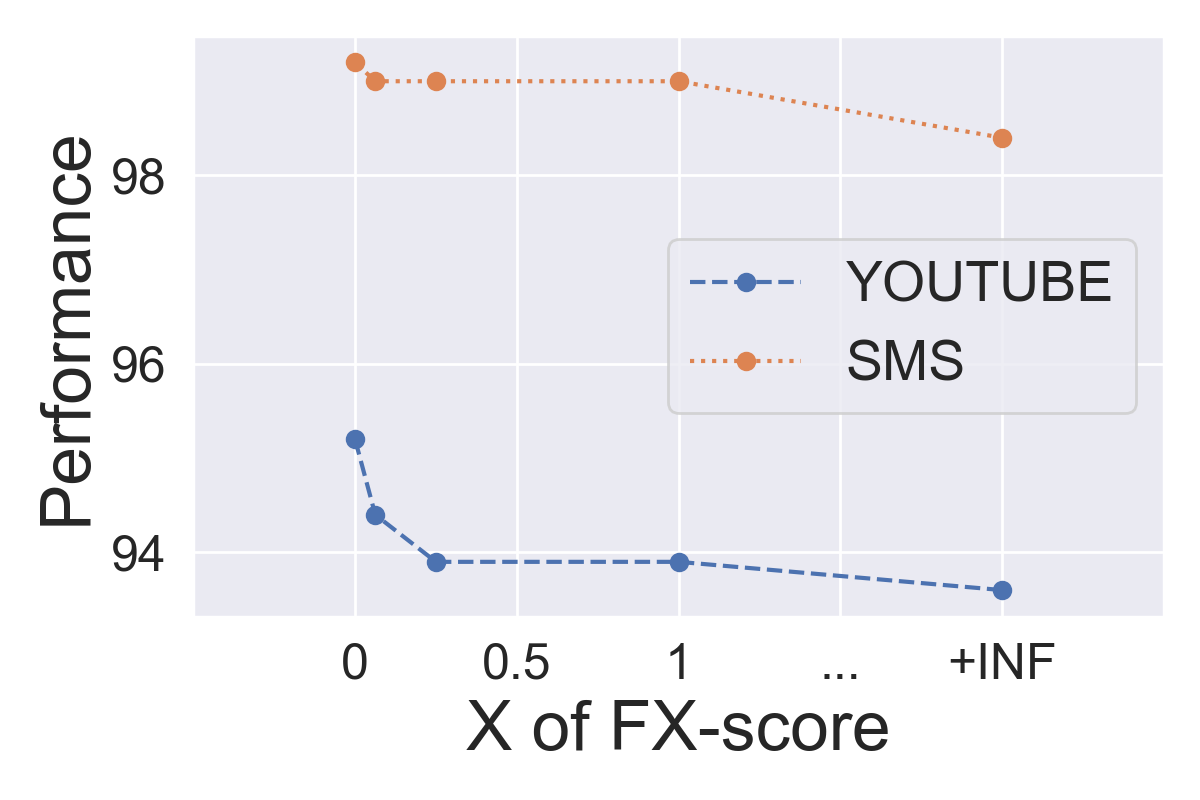}
\caption{The influence of X of FX-score as the selection threshold on the final performance.}
\label{fig:fxscore}
\end{figure}

In practice, we find that choosing a proper threshold for selection model might have some influence on the final performance. In detail, we choose FX-score as the target to choose the threshold which yields best FX-score on the dev\_eval set, and use this threshold to select the samples from the unlabeled data based on the output of the selection model. We investigate the influence on final performance by changing the X of FX-score on YOUTUBE and SMS. The computation of this metric is displayed as follow:

\begin{equation}
\label{equ1}
FX-score = \frac{(1+X^2)*precision*recall}{X^2*preision+recall}
\end{equation}
Noted that this metric becomes F1-score if we set X as 1. The X measures the preference of precision to recall. If X approaches 0, it becomes precision. If X approaches infinite, it becomes recall. As shown in Figure~\ref{fig:fxscore}, the performance gradually decreases as the X grows, which implicitly indicates that precision matters more than recall when we are going to select samples from unlabeled data.

\subsubsection{The Performance Under The Few-shot Setting}
\begin{figure}[h]
\centering
\subfigure[YOUTUBE]{
\label{fig:few_youtube}
\includegraphics[width=0.8\linewidth]{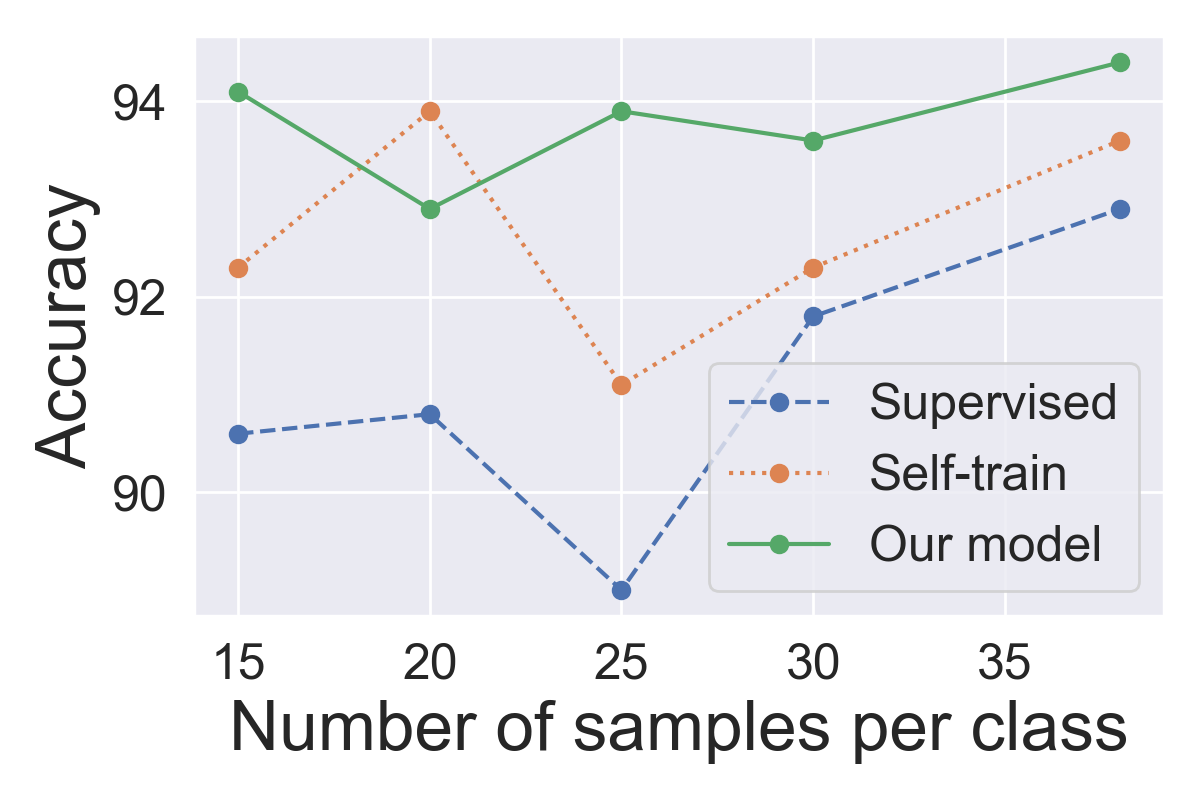}}
\\
\subfigure[IMDB]{
\label{fig:few_imdb}
\includegraphics[width=0.8\linewidth]{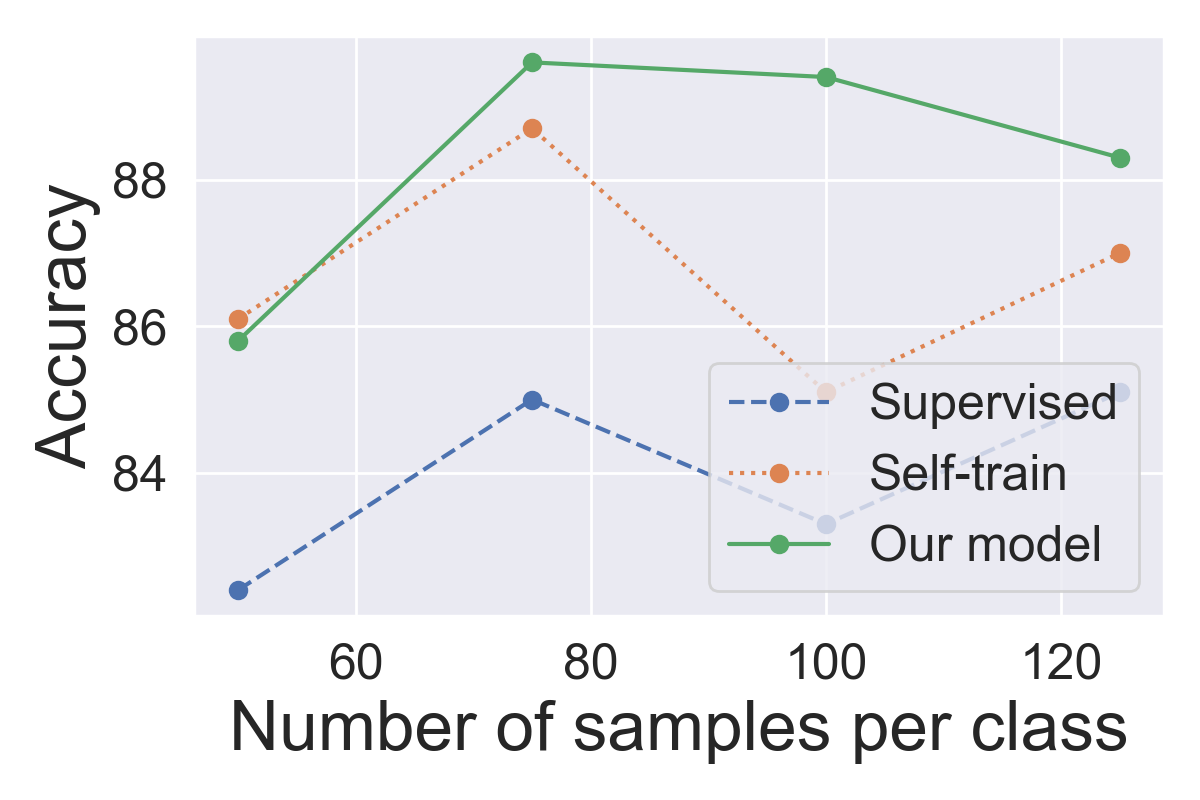}}
\caption{Evaluation on Youtube, SMS and IMDB under the few-shot setting.}
\label{fig:few_shot}
\end{figure}

We also investigate the performance of IMDB and YOUTUBE under the few-shot learning setting. The results are shown in Figure~\ref{fig:few_shot}.
\label{sec:few_shot}
\subsubsection{Case Study: Signal Difference On Train and Development Sets.}
\begin{figure}[h]
\tiny
\centering
\includegraphics[width=1\linewidth]{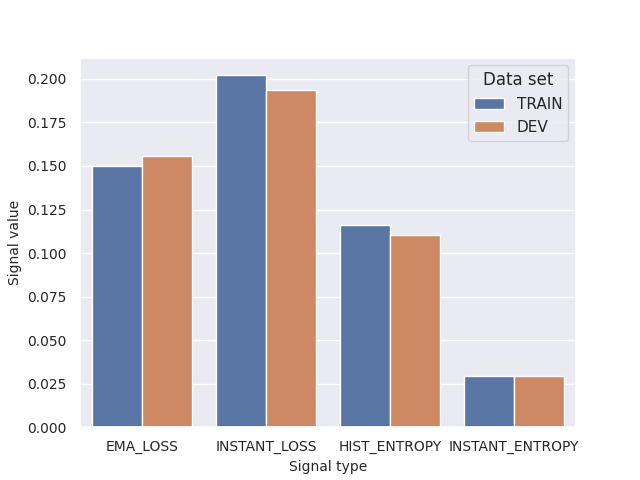}
\caption{Mean of signals on hard level training and development set}
\label{fig:worse_histgram}
\end{figure}

\begin{figure}[h]
\tiny
\centering
\includegraphics[width=1\linewidth]{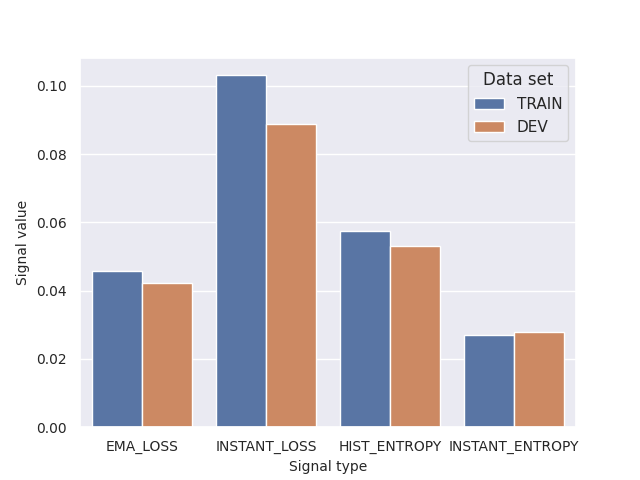}
\caption{Mean of signals on medium level training and development set}
\label{fig:normal_histgram}
\end{figure}

\begin{figure}[h]
\tiny
\centering
\includegraphics[width=1\linewidth]{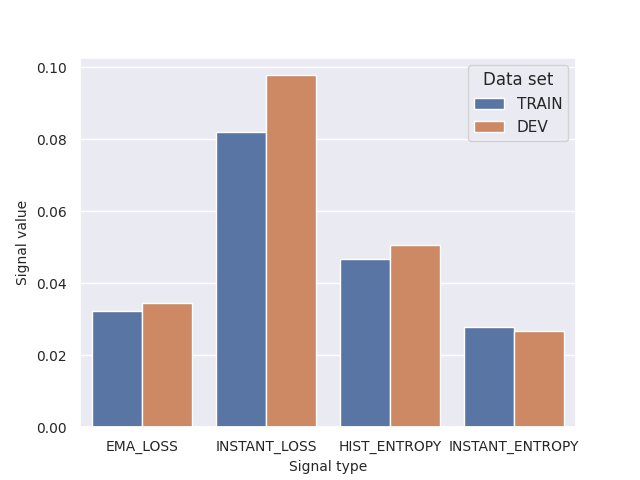}
\caption{Mean of signals on easy level training and development set}
\label{fig:better_histgram}
\end{figure}

We show the difference of signals on training and development sets in Figure~\ref{fig:worse_histgram},~\ref{fig:normal_histgram},~\ref{fig:better_histgram}. We can see that the all signals show close statistics on both sets. This indicates that our noise transfer approach holds and has a good performance. This phenomena exists in all three error levels.

\subsubsection{Influence of Signals On AISHELL-1 Dev}

We show the influence of different signals on AISHELL-1 development set in Table~\ref{tab:signalabalationonasrfull}. We can see that the tendency is same as the tendency on the test set.

\begin{table}[h]
% \small
  \centering
    \begin{tabular}{c|ccc}
    \midrule
    Signal & Hard  & Medium & Easy  \\
    \midrule
    \midrule
        All & \textbf{23.97}  & \textbf{16.01}  & \textbf{11.96}     \\
            -EMAL      & 24.35  & 16.43  & 12.45 \\
    -IL        & 25.48  & 17.55  & 13.71  \\

    -HE            & 25.89  & 18.01  & 13.99\\

    \bottomrule
    \end{tabular}%
    \caption{Ablation experiments for signals on dev set of AISHELL-1. Each line is removing one signal from previous line. }
    \label{tab:signalabalationonasrfull}%
\end{table}%

\end{document}